\documentclass[11pt]{article}

\usepackage[final]{acl}

\usepackage{times}
\usepackage{latexsym}

\usepackage[T1]{fontenc}

\usepackage[utf8]{inputenc}

\usepackage{microtype}

\usepackage{inconsolata}

\usepackage{graphicx}

\usepackage{amsmath}
\usepackage{amsthm} 
\theoremstyle{definition} 
\newtheorem{definition}{Definition}
\usepackage{booktabs}
\usepackage{listings}
\usepackage{float}
\usepackage{hyperref}

\title{Scoped Semantic Invariance as a Foundation for LLM Reliability}

\author{{\bf Rosario Uceda-Sosa} \\ {\bf Maria Chang}  \\ 
    {\bf Karthikeyan Natesan Ramamurthy} \\  {\bf Moninder Singh} \\
  IBM T. J. Watson Research Center, New York, USA \\
  }

\begin{document}
\maketitle
\begin{abstract}

LLMs often answer semantically equivalent queries inconsistently, even when the underlying facts are known. We formalize consistency as \emph{semantic invariance} over equivalence classes of queries and show that invariance under arbitrary meaning-preserving transformations is undecidable in general. This motivates \emph{scoped semantic invariance}, where model behavior is evaluated relative to bounded semantic domains.

We operationalize this framework using knowledge graphs (KGs), which act as explicit semantic control structures defining local semantics and grounding contexts. Across the MITRE D3FEND ontology and eight Wikidata-derived domains, we show that inconsistencies are structured rather than random, concentrating in localized hotspot subgraphs organized around recurring TREE, STAR, and CYCLE patterns that remain stable across models and domains.

We further show that selective graph grounding improves consistency with the KG specification and cross-model agreement, while indiscriminate grounding introduces regressions. A layered grounding strategy achieves the strongest monotonicity profile, preserving KG-consistent knowledge while reducing inconsistency. Finally, we show that hotspot structures discovered from simple consistency probes transfer to more realistic compositional reasoning tasks, positioning them as a practical reliability layer for predicting and targeting failures rather than a canonical decomposition.

\end{abstract}
\section{Introduction}

Large language models (LLMs) achieve strong performance on many benchmarks, yet provide no guarantees of consistency across paraphrases or simple reasoning variations. Prior work shows that models often answer queries about knowledge graph (KG) relations in a manner consistent with it, yet fail on semantically equivalent formulations~\cite{Elazar2021}. Thus, even when models \emph{know} the underlying facts, their behavior may remain semantically unstable.

This instability reflects a broader limitation of current LLM interaction paradigms. Prompt engineering mixes instructions, data, and control flow within unstructured natural language, making behavior brittle, difficult to reproduce and monitor across domains and models~\citep{LLMon_1}. Our goal is not to eliminate prompting, but to make it more disciplined by grounding interactions in explicit semantic structure. We view KGs as a step toward such a semantic interface: a structured substrate against which model behavior can be evaluated, monitored, and selectively constrained. Our aim throughout is not to maximize aggregate accuracy, but to make failures predictable and localized enough that they can be selectively identified and repaired — a shift in emphasis from average-case performance to targeted reliability interventions.

A natural reliability requirement in this setting is \emph{semantic invariance}: meaning-preserving queries should yield identical outputs~\citep{Semantic_Invariance}. Recent work similarly identifies consistency and robustness as central dimensions of LLM reliability~\citep{AI_Agent_Reliability}. However, semantic invariance cannot be guaranteed in general: for unconstrained tasks, consistency under semantic transformations is undecidable~\citep{sipser13, Abiteboul1995}. This motivates a shift from global guarantees to \emph{scoped semantic invariance}, where model behavior is evaluated relative to bounded semantic domains.

We operationalize scoped invariance using KGs. A KG encodes explicit semantic structure through relations such as hierarchy, transitivity, and property inheritance, inducing equivalence classes of semantically related queries, which we refer to as \textit{query clusters}. This makes consistency measurable relative to a well-defined semantic context while exposing the local constraints governing reasoning consistent with the KG specification.

Within this setting, we find that consistency failures are highly structured. Rather than being uniformly distributed, they concentrate in localized \emph{hotspot subgraphs} that recur across models and domains. These hotspots decompose into a small set of recurring structural patterns---TREE, STAR, and CYCLE---which predict where models are likely to behave inconsistently. This suggests that semantic instability is shaped primarily by KG topology rather than incidental prompt variation.

\begin{figure*}
  \centering
\includegraphics[width=\linewidth]{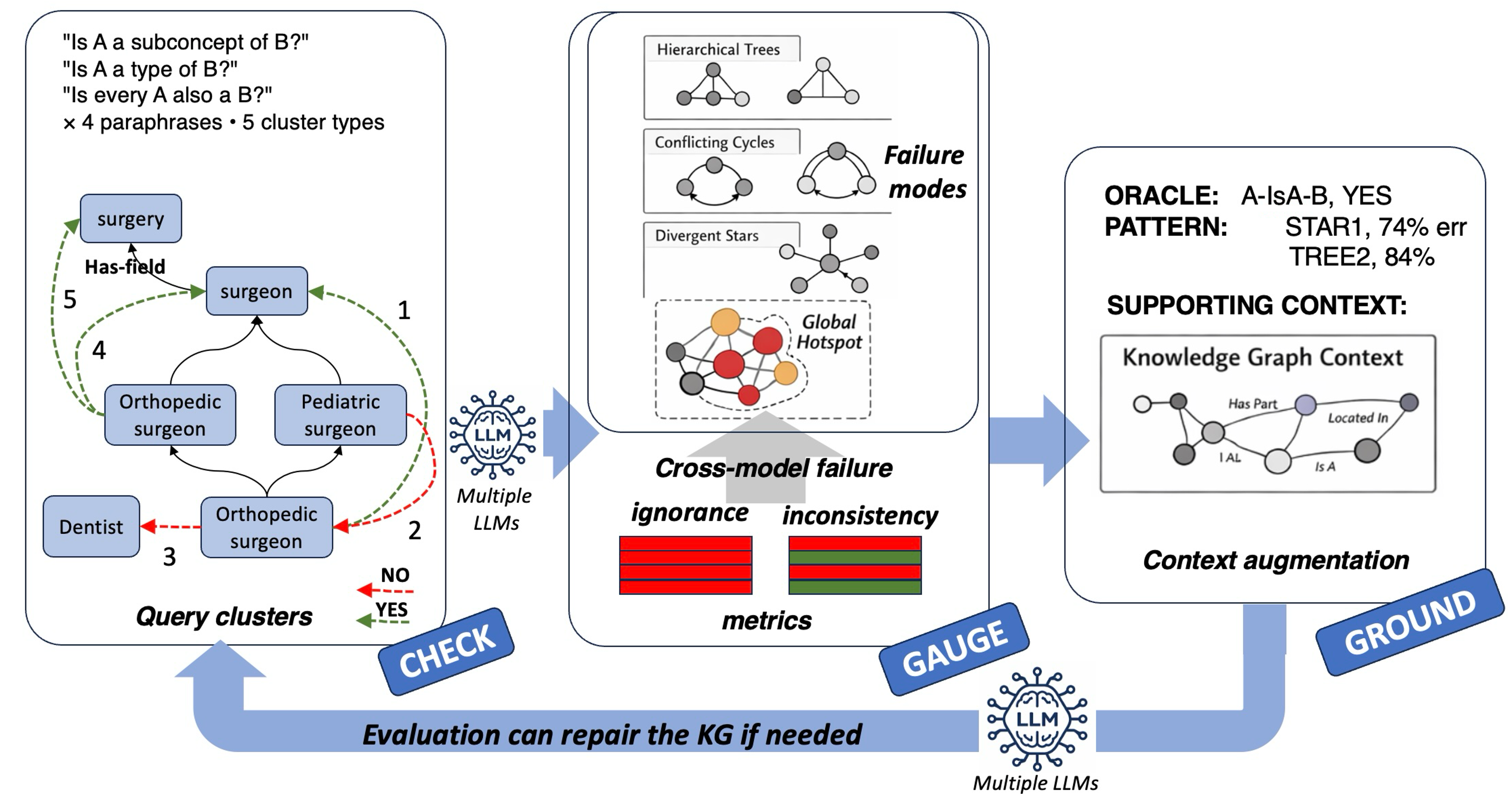}
  \caption{Pipeline overview. CHECK establishes baseline consistency and accuracy. GAUGE identifies hotspot subgraphs and structural failure patterns. GROUND applies selective semantic grounding strategies.}
  \label{pipeline}
\end{figure*}

Building on this structure, we study selective graph grounding through KG augmentation. We show that indiscriminate augmentation can improve aggregate performance while simultaneously introducing local regressions. This motivates \emph{monotonicity} as a reliability criterion: interventions should reduce inconsistency while also preserving KG-aligned knowledge. We therefore use a layered strategy that applies stronger structural grounding only where simpler grounding fails.

Our primary goal is not unrestricted repair, but principled semantic evaluation and control over bounded domains. In practical deployments, KGs provide an explicit semantic reference against which LLM behavior can be audited, benchmarked, and selectively constrained, particularly in industrial or mission-critical settings where instability over a small set of concepts may have disproportionate operational impact.

We evaluate these ideas on a real-world cybersecurity dataset, \textsc{SI-Cyber}, derived from the MITRE D3FEND ontology~\citep{mitre_d3fend_ontology}, together with multiple Wikidata-derived domains~\citep{Uceda_2024}. 

As preliminary evidence that these patterns generalize beyond controlled equivalence-cluster probes, we further observe, on a small set (N=25) of human-authored compositional reasoning queries (Section~\ref{appendix:testing_realistic_queries}), that selective graph grounding from simple queries improves consistency and cross-model agreement; we treat this as a targeted confirmatory signal rather than a large-scale validation of the taxonomy.

\noindent\textbf{Contributions.}
(1) We formalize consistency as invariance over KG-derived equivalence classes and establish fundamental limits on semantic invariance.
(2) We introduce \textsc{SI-Cyber}, a cybersecurity benchmark for scoped semantic invariance\footnote{\url{https://huggingface.co/datasets/ibm-research/SI-Cyber}}.
(3) We show empirically that consistency failures are structured and predictive, concentrating in recurring hotspot patterns across models and domains.
(4) We show that every hotspot edge receives a CYCLE, STAR, or TREE type under a deterministic classification procedure, with TREE as the default case.
(5) We show that KG augmentation improves aggregate performance but may introduce local regressions, motivating monotonicity and layered grounding as practical reliability constraints.
\section{Scoped Semantic Invariance}
\label{sec:scoped_semantic_invariance}

A natural reliability requirement for LLMs is \emph{semantic invariance}: meaning-preserving queries should produce identical outputs. However, this property cannot be guaranteed in general for open-domain models (formal treatment in Appendix~\ref{appendix:undecidability}).

At a high level, guaranteeing semantic invariance requires deciding whether two inputs induce the same semantic behavior under the model. For sufficiently expressive query spaces, this reduces to a non-trivial semantic equivalence problem and is therefore undecidable~\citep{sipser13, Abiteboul1995}. Intuitively, general-purpose LLMs operate over an unbounded semantic domain in which logically equivalent queries may still activate distinct internal representations due to contextual, distributional, or usage-dependent distinctions learned during training. We provide a formal treatment in Appendix~\ref{appendix:undecidability}.

This limitation motivates a shift from global to \emph{scoped} semantic invariance. Rather than attempting to enforce consistency universally, we evaluate model behavior relative to a bounded semantic domain defined by a knowledge graph (KG).

Let $\mathcal{E}$ be a finite set of equivalence classes induced by a KG, where each class contains semantically related queries (e.g., paraphrases, transitive implications, or logically equivalent formulations of a KG relation). A model $F$ satisfies scoped semantic invariance over $\mathcal{E}$ if:
\[
\forall E \in \mathcal{E}, \quad \forall q_1,q_2 \in E,\quad F(q_1)=F(q_2).
\]

Under this formulation, consistency becomes measurable relative to an explicit semantic structure rather than an unrestricted input space. KGs are particularly suitable for this role because they encode hierarchy, transitivity, and property inheritance explicitly, exposing the local semantic constraints governing reasoning consistent with the KG.

The key question then becomes empirical rather than purely theoretical: when semantic invariance is scoped to a KG, are failures random, or do they exhibit underlying structure?
\subsection{Hotspot Subgraph Patterns: The Structure of Inconsistency}
\label{sec:hotspot_patterns}

Violations of scoped semantic invariance are not uniformly distributed. Instead, they concentrate in localized regions of the KG, which we call \emph{hotspot subgraphs}.

A \emph{hot edge} is a KG edge whose associated query cluster falls below predefined consistency or accuracy thresholds. A \emph{hotspot subgraph} is a connected region induced by such edges across models.

These subgraphs decompose into recurring structural \emph{patterns}~\citep{milo2002network} that provide an interpretable bridge between graph topology and reasoning failures. We decompose each connected hotspot component by a deterministic procedure: cycle edges are identified first — a self-loop, or a directed cycle in which every participating node has undirected degree exactly two — and extracted as CYCLE patterns; the remaining acyclic edges are then partitioned into STAR and TREE patterns, where a subgraph is STAR only if a single hub node is the unique articulation point separating every other node pair, and TREE otherwise (the default case, covering single edges, chains, multi-hub or cycle-sharing structures such as two cycles sharing an edge, and dense subgraphs such as K4). A single hotspot component may thus yield more than one pattern — e.g., a cycle plus an attached branch — but every hotspot edge receives exactly one pattern type, without claiming a canonical graph-theoretic decomposition.

\begin{figure*}
  \centering
\includegraphics[width=\linewidth]{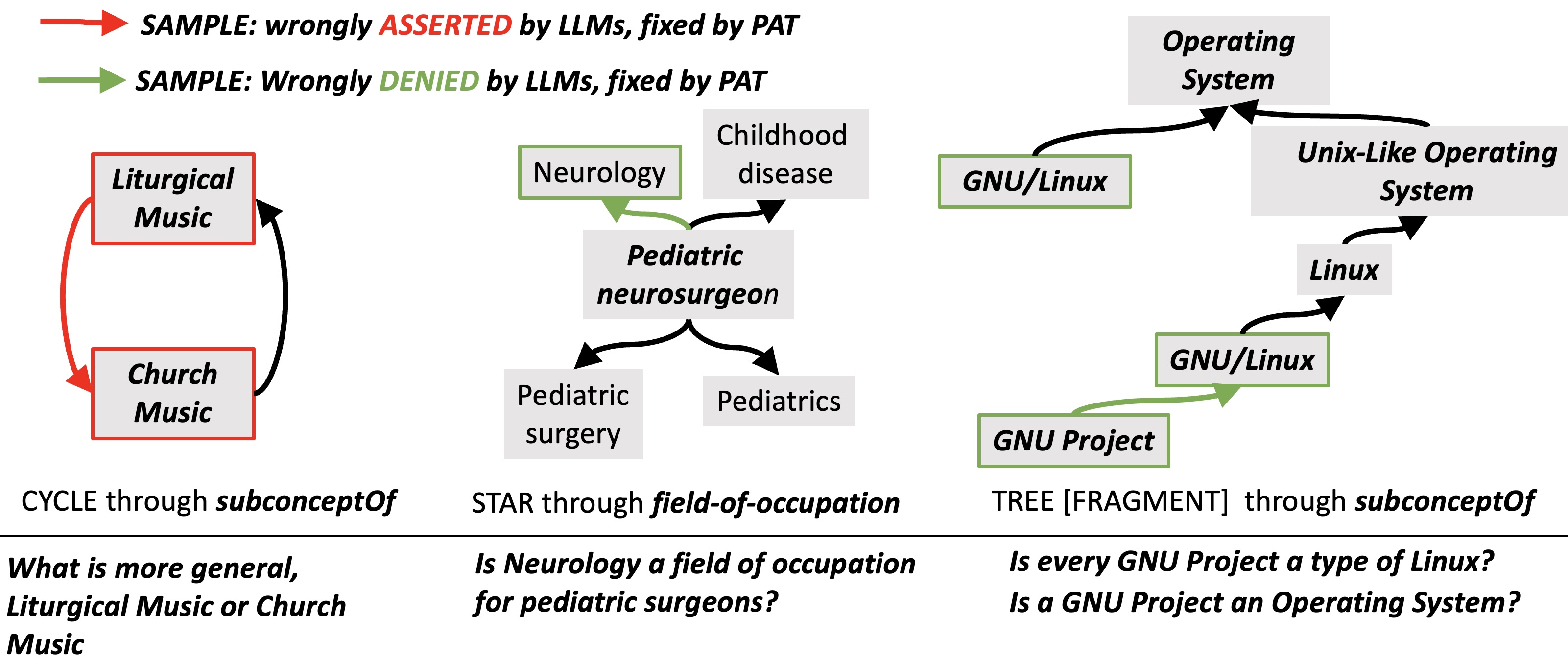}
  \caption{Examples of hotspot subgraph patterns. Red edges denote inconsistent relations accepted by the model; green edges denote corrective structural grounding derived from the KG.}
  \label{fig:pattern_samples}
\end{figure*}

\paragraph{CYCLE.}
CYCLE patterns capture \emph{conceptual collapse}: models incorrectly treating nearby concepts as equivalent. In Figure~\ref{fig:pattern_samples}, the KG asserts that \emph{church music} is a subconcept of \emph{liturgical music}, but not the reverse. Some models nonetheless accept the inverse relation, effectively behaving as if the two concepts were interchangeable and thereby inducing a cycle that does not exist in the underlying ontology.

Although true cycles are rare in curated KGs, hotspot cycles still emerge because models may collapse nearby concepts into equivalent internal representations. The prevalence of this failure mode varies sharply by domain — from under 3\% of patterns in SI-Cyber and music genres to over 20\% in natural disasters and nearly 38\% in home appliances — suggesting that conceptual collapse is not a fixed base rate but depends on how densely a domain's concepts cluster around near-synonymous distinctions.

\paragraph{STAR.}
STAR patterns capture failures of \emph{generalization across related concepts}. In Figure~\ref{fig:pattern_samples}, some models correctly identify that a neurologist specializes in neurology, yet fail to extend this reasoning to a pediatric neurosurgeon, effectively treating the pediatric specialization as overriding the neurological one. The shared hub structure exposes the missing semantic dependency.

\paragraph{TREE.}
TREE patterns capture failures of \emph{transitive reasoning and inheritance} along \texttt{IsA} hierarchies. In Figure~\ref{fig:pattern_samples}, Wikidata places GNU Project (Q7598) under Linux (Q388) in the subconcept hierarchy. Models often resist this relation because they distinguish GNU Project (the software initiative) from Linux (the kernel), despite recognizing their close association. The tree structure exposes the transitive semantic commitments implied by the KG.

It is worth noting that we do not claim that the KG is necessarily correct. Rather, we use it as a controllable semantic structure against which model behavior can be evaluated and monitored, particularly in mission-critical settings. 

The existence of localized hotspot patterns suggests that semantic inconsistency is structured rather than random, motivating graph-grounded evaluation and selective intervention strategies.
\section{Selective Graph Grounding}
\label{pipeline_description}

The potential for violations of scoped semantic invariance to concentrate in localized hotespot patterns suggests a structured evalation strategy: measure consistency relative to KG structure, identify unstable semantic regions, and selectively ground model behavior within those regions.

We operationalize this idea through a three-stage pipeline (Figure~\ref{pipeline}) over KG-derived query clusters:

\begin{enumerate}
    \item \textbf{CHECK.} Evaluate baseline consistency and accuracy over semantically equivalent query clusters without additional grounding.

    \item \textbf{GAUGE.} Identify hotspot subgraphs by locating edges and query clusters exhibiting low consistency or accuracy across models.

    \item \textbf{GROUND.} Re-evaluate unstable regions using selective graph grounding derived from the KG.
\end{enumerate}

We study the following grounding conditions:

\begin{itemize}
    \item \textbf{BL (Baseline):} no KG grounding.
    
    \item \textbf{VO (Verdict-Only):} inject the target KG relation as an explicit semantic constraint. VO is intentionally unrealistic and serves as an oracle upper bound, since it assumes direct access to the queried KG edge.
    
    \item \textbf{PAT (Pattern-Aware):} augment VO with local structural context derived from hotspot patterns (TREE, STAR, or CYCLE).

    \item \textbf{NRAG (Naive RAG):} where the document set consists of the manually curated natural language (English) comments associated with each edge in the hotspot pattern.
    
    \item \textbf{LAY (Layered):} apply stronger structural grounding only when simpler grounding fails.
\end{itemize}

A central reliability requirement is \emph{monotonicity}: grounding should improve or preserve consistency with the KG specification without introducing regressions. In practice, indiscriminate augmentation may improve aggregate metrics while simultaneously breaking query clusters that were KG-aligned at baseline. The layered strategy addresses this by restricting stronger structural grounding to unstable regions only.

Importantly, the KG does not function merely as external knowledge retrieval. Instead, it provides explicit semantic constraints that expose the local structure governing reasoning consistent with the KG specification. Additional implementation details and full prompting configurations are provided in Appendix~\ref{appendix:extended_results}.

\section{Experimental Setup}
\label{sec:experimental_setup}

\paragraph{Datasets.}
We construct \textsc{SI-Cyber} (provided in submission), a dataset of 7,980 clusters derived from MITRE D3FEND~\citep{mitre_d3fend_ontology,d3fend_resources}, a cybersecurity ontology encoding defensive techniques, system artifacts, and adversarial behaviors. Unlike general-purpose KGs, D3FEND contains specialized ontological distinctions that often diverge from common intuition, providing a challenging setting for semantic consistency evaluation. Full dataset construction details and SPARQL extraction queries are provided in Appendix~\ref{appendix:dataset_construction}.

We additionally evaluate eight Wikidata-derived domains introduced in~\citep{Uceda_2024,Uceda_2025}, extended here with explicit target-concept metadata to support hotspot subgraph computation. These domains complement the specialized semantic structure of D3FEND. To assess whether hotspot patterns transfer beyond controlled equivalence clusters, we also include a set of 25 manually authored realistic queries (listed in Appendix~\ref{appendix:complex_queries} and analyzed in ~\ref{appendix:testing_realistic_queries}) which span several of these Wikidata domains.

\paragraph{Models.}
We evaluate six LLMs spanning multiple architectures and parameter scales: Llama-3.3-70B-Instruct~\citep{llama3}, Qwen3.5-397B-A17B-FP8~\citep{qwen}, gpt-oss-120b~\citep{gpt_oss}, DeepSeek-V3.2~\citep{deepseek}, Granite-4.0-h-small~\citep{granite}, and Mistral-Small-3.2-24B-Instruct~\citep{mistral}. Additional implementation details are provided in Appendix~\ref{appendix:prompting_protocol}.

\paragraph{Metrics.}
We evaluate both semantic correctness and semantic invariance. Metrics are computed for a single run for each condition.

\emph{Balanced accuracy}~\citep{brodersen2010balanced} measures correctness under class imbalance:
\[
\mathrm{BalAcc} =
\frac{\mathrm{TPR}+\mathrm{TNR}}{2}.
\]

 An edge is flagged hot for a model if it fails any of five criteria: accuracy < 0.80, consistency < 0.80, conflicting majority votes across paraphrases, balanced accuracy < 0.75, or specificity < 0.80; an edge is a hotspot if it is flagged hot by at least 2 of the 6 models.We use a hotspot threshold of 0.80 to capture asymmetric failure modes. This penalizes directionally biased behavior (e.g., consistently answering ``yes''), which may still achieve high raw accuracy under class imbalance.

 To confirm these results are not an artifact of a specific cutoff, we swept the accuracy/consistency/specificity criteria over {0.75, 0.80, 0.85} (with balanced accuracy scaled correspondingly). The hotspot edge set is stable, with Jaccard overlap $\geq$ 98.8\% against the 0.80 operating point across all but the loosest cut (86.2\% at 0.75/0.75/0.75).

\emph{Consistency} measures invariance across paraphrases within a query cluster:
\[
\mathrm{Consistency} =
\frac{
|\{E \in \mathcal{E} : \mathrm{Consistent}(E)=1\}|
}{
|\mathcal{E}|
}
\]

Because single-edge clusters contain four paraphrased queries, we use a consistency threshold greater than 0.75 to detect even minimal paraphrase instability.

Aggregate metrics alone are insufficient to evaluate intervention reliability since augmentation may improve overall performance while degrading previously KG-aligned knowledge. Hence, we additionally evaluate monotonicity under intervention.

\emph{Knowledge Retention Rate (KRR)} measures preservation of baseline-consistent knowledge:
\[
\mathrm{KRR} =
\frac{
\mathrm{baseline\_correct}-\mathrm{broken}
}{
\mathrm{baseline\_correct}
}.
\]

\emph{Fix/Break Ratio (F/B)} measures intervention efficiency as the ratio between corrected and newly broken clusters.

Together, KRR and F/B evaluate whether grounding improves consistency with the KG-specification without introducing regressions. A repeated-run check with temperature 0.3 (vs. 0 in the main results) confirms these results are not an artifact of decoding-randomness averaging (see Appendix~\ref{sec:sensitivity_to_randomness} for details)

\section{Key Findings}
\label{sec:results}

Across domains and models, aggregate baseline performance is high (Table~\ref{tab:baacc_consistency_summary}), yet semantic consistency remains imperfect. VO and PAT assume oracle access to the queried KG edge and therefore serve as diagnostic upper bounds. NRAG is the deployable condition: it uses only natural-language structural context and matches PAT's balanced accuracy (96.871\%) without oracle access. LAY is included to study selective intervention and monotonicity, but is not directly deployable because it incorporates an oracle component.

\begin{table*}[t]
\centering
\small
\begin{tabular}{lrrrr}
\toprule
\textbf{Approach} &
\textbf{Bal Acc (\%)} &
\textbf{Bal Acc $\Delta$ from BL} &
\textbf{Consistency (\%)} &
\textbf{Consistency $\Delta$ from BL} \\
\midrule
BL   & 86.469 & 0.000  & 95.350 & 0.000 \\
NRAG & 96.871 & 10.402 & 97.994 & 2.644 \\
VO   & 96.659 & 10.190 & 97.888 & 2.538 \\
PAT  & 96.871 & 10.402 & 97.994 & 2.644 \\
\textbf{LAY}  
     & \textbf{96.935} 
     & \textbf{10.466} 
     & \textbf{98.008} 
     & \textbf{2.657} \\
\bottomrule
\end{tabular}
\caption{Balanced accuracy and consistency results pooled {\em across clusters} from Baseline (BL), Naive Rag (NRAG), Verdict-Only (VO), Patterns (PAT) and Layered (LAY).
}
\label{tab:baacc_consistency_summary}
\end{table*}

\paragraph{Failures are structured and largely model-invariant.}
Across nine domains (Table~\ref{tab:hotspot_per_domain}), hotspot patterns are dominated on average by \textbf{STAR} (49.81\%) and \textbf{TREE} (36.93\%), with \textbf{CYCLE} less frequent overall (13.25\%). The structural mix varies substantially by domain: CYCLE reaches 21.74\% in natural disasters and 37.93\% in home appliances, while SI-Cyber is strongly TREE-dominated (79.19\%). Hotspot density ranges from 3.37\% (SI-Cyber) to 17.57\% (finance), typically around 10--17\%, indicating that failures are localized rather than uniformly distributed across the KG.


\begin{table*}[t]
\centering
\small
\setlength{\tabcolsep}{4pt}
\begin{tabular}{lrrrrrrr}
\toprule
\textbf{Domain} &
\textbf{Patterns} &
\textbf{\% STAR} &
\textbf{\% TREE} &
\textbf{\% CYCLE} &
\textbf{Hot Edges} &
\textbf{Total Edges} &
\textbf{Density} \\
\midrule
academic disciplines & 20  & 70.00 & 25.00 & 5.00  & 49  & 411  & 11.92 \\
dishes                & 43  & 51.16 & 34.88 & 13.95 & 128 & 1038 & 12.33 \\
finance               & 35  & 65.71 & 25.71 & 8.57  & 104 & 592  & 17.57 \\
home appliances       & 29  & 41.38 & 20.69 & 37.93 & 71  & 491  & 14.46 \\
medical specialties   & 34  & 61.76 & 20.59 & 17.65 & 117 & 706  & 16.57 \\
music genres          & 44  & 50.00 & 47.73 & 2.27  & 200 & 1921 & 10.41 \\
natural disasters     & 23  & 34.78 & 43.48 & 21.74 & 52  & 305  & 17.05 \\
software              & 37  & 54.05 & 35.14 & 10.81 & 200 & 1375 & 14.55 \\
\textbf{SI-Cyber} &
\textbf{149} & \textbf{19.46} & \textbf{79.19} & \textbf{1.34} &
\textbf{200} & \textbf{5935} & \textbf{3.37} \\
\textbf{AVG} &
\textbf{46} & \textbf{49.81} & \textbf{36.93} & \textbf{13.25} &
\textbf{125} & \textbf{1419} & \textbf{13.14} \\
\bottomrule
\end{tabular}
\caption{Hotspot subgraph patterns obtained from the baseline (BL) across domains.}
\label{tab:hotspot_per_domain}
\end{table*}

Because every hotspot edge belongs to a connected hotspot component, and every connected component admits a TREE, STAR, or CYCLE assignment (with single edges and chains treated as degenerate trees), these statistics characterize the global structural organization of failures rather than a sampled subset.

Across models (Table~\ref{tab:hotspot_per_model}), hotspot density varies substantially (13.4\% for Llama vs.\ 26.1\% for Mistral), reflecting differences in overall model quality. However, the distribution of hotspot patterns remains remarkably stable: approximately 52\% STAR, 44\% TREE, and 3--5\% CYCLE across all models. Model strength therefore affects primarily \emph{how many} edges fail, not \emph{which structural configurations} fail.

\begin{table*}[t]
\centering
\small
\setlength{\tabcolsep}{4pt}
\begin{tabular}{lrrrrrrr}
\toprule
\textbf{Model} &
\textbf{Patt's} &
\textbf{\% STAR} &
\textbf{\% TREE} &
\textbf{\% CYCLE} &
\textbf{Hot Edges} &
\textbf{\# Edges} &
\textbf{Density} \\
\midrule
DeepSeek-V3.2          & 1574 & 53.30 & 43.70 & 3.00 & 2723 & 12774 & 21.30 \\
Mistral-Small-3.2-24B  & 1751 & 52.90 & 44.00 & 3.00 & 3330 & 12774 & 26.10 \\
Qwen3.5                & 1507 & 50.80 & 45.40 & 3.80 & 2402 & 12774 & 18.80 \\
gpt-oss-120b           & 1278 & 53.10 & 43.70 & 3.20 & 2208 & 12774 & 17.30 \\
granite-4.0-h-small    & 1571 & 52.30 & 44.40 & 3.20 & 2770 & 12774 & 21.70 \\
llama-3-3-70b-instruct & 1108 & 50.70 & 44.30 & 5.00 & 1706 & 12774 & 13.40 \\
\textbf{AVG}           & \textbf{1464} & \textbf{52.18} & \textbf{44.25} & \textbf{3.53} & \textbf{2523} & \textbf{12774} & \textbf{19.76} \\
\bottomrule
\end{tabular}
\caption{Hotspot subgraph patterns after running the baseline across models.}
\label{tab:hotspot_per_model}
\end{table*}

\paragraph{Hotspot status is predictive across models, not merely descriptive.}
A natural concern is that hotspot patterns could be artifacts of a single model's idiosyncrasies rather than genuine structural signal. To test this, we run a leave-one-model-out analysis: for each held-out model, we label an edge as predicted-hot if it is hot for at least one of the remaining five models, and check whether this predicts hotness on the held-out model, now using the complete five-criterion hotness rule (accuracy, consistency, conflicting majorities, balanced accuracy, specificity). Pooled across all edges and held-out models~\ref{tab:leave_one_model_out}, hotspot status predicts held-out hotness with 76.4\% precision against a 27.0\% base rate (2.87× lift) and a pooled AUC of 0.905, with the effect holding individually for every held-out model (57–96\% precision, lift 2.4–3.1×). 

To rule out the possibility that this signal merely reflects structurally obvious properties, we compare against two baselines requiring no cross-model information: edge degree (source+target connectivity) and query frequency (mean questions per edge). Cross-model agreement (AUC 0.905) substantially outperforms both frequency (AUC 0.637) and degree (AUC 0.534, indistinguishable from chance).This indicates that hotspot subgraphs identify structurally unstable regions of the ontology that generalize across model architectures, rather than model-specific noise — the predictive signal a purely descriptive, definition-driven decomposition would not produce.

\begin{table*}[t]
\centering
\small
\setlength{\tabcolsep}{4pt}
\begin{tabular}{lrrrrrr}
\toprule
\textbf{Held-out Model} &
\textbf{\# Edges} &
\textbf{Base} &
\textbf{Prec.} &
\textbf{Recall} &
\textbf{Acc.} &
\textbf{Lift} \\
\midrule
Qwen3.5-397B-A17B-FP8   & 12774 & 0.239 & 0.733 & 0.838 & 0.888 & 3.06 \\
DeepSeek-V3.2            & 12774 & 0.285 & 0.860 & 0.768 & 0.899 & 3.02 \\
granite-4.0-h-small      & 12774 & 0.302 & 0.838 & 0.704 & 0.869 & 2.77 \\
llama-3-3-70b-instruct   & 12774 & 0.194 & 0.566 & 0.808 & 0.843 & 2.92 \\
Mistral-Small-3.2-24B    & 12774 & 0.392 & 0.955 & 0.581 & 0.825 & 2.43 \\
gpt-oss-120b             & 12774 & 0.208 & 0.630 & 0.836 & 0.863 & 3.02 \\
\textbf{Mean}            & \textbf{13855} & \textbf{0.270} & \textbf{0.764} &
\textbf{0.756} & \textbf{0.865} & \textbf{2.87} \\
\midrule
\multicolumn{7}{l}{\textbf{Pooled AUC: 0.905}} \\
\bottomrule
\end{tabular}
\caption{Leave-one-model-out prediction of edge hotness. For each held-out model,
an edge is predicted hot if it is hot for $\geq 1$ of the other five models;
Base is the held-out model's true hot rate.}
\label{tab:leave_one_model_out}
\end{table*}

\paragraph{Hotspot structure enables lightweight semantic control.}
Hotspot subgraphs are sparse (13\% mean domain density, Table~\ref{tab:hotspot_per_domain}) and consist of small, informative edge sets rather than dense neighborhoods. Because these regions organize into recurring TREE, STAR, and CYCLE structures, grounding reduces to lightweight structural lookup over small local contexts (typically 5--8 edges), suggesting that KG grounding can function as a lightweight semantic control layer rather than a large-scale retrieval system.

\paragraph{Reliable grounding requires monotonicity.}
KG grounding improves aggregate performance (Table~\ref{tab:baacc_consistency_summary}), increasing both balanced accuracy and consistency across all augmentation methods. However, non-layered strategies also introduce regressions, breaking clusters that were KG-consistent at baseline (Tables~\ref{tab:knowledge_stability_cluster} and~\ref{tab:knowledge_stability_query}). Aggregate gains alone are therefore insufficient to characterize reliability.

Across both cluster- and query-level evaluations, the layered strategy (\textbf{LAY}) achieves the strongest stability profile: highest fix rate, lowest break rate, and best KRR. PAT and NRAG achieve similar grounding performance, but PAT does so with substantially lower token cost. VO is weak as a standalone strategy, yet effective within LAY as a lightweight first-pass grounding condition.

Query-level results are consistently more favorable than cluster-level ones, indicating that failures are spatially concentrated rather than randomly scattered. Structured grounding therefore tends to ground multiple queries within unstable regions while breaking only isolated queries in otherwise KG-consistent clusters.

\begin{table*}[t]
\centering
\small
\setlength{\tabcolsep}{4pt}
\begin{tabular}{lrrrrrrrr}
\toprule
\textbf{Approach} &
\textbf{Avg Tokens} &
\textbf{Avg Triples} &
\textbf{Fixed} &
\textbf{\%Fixed} &
\textbf{Broken} &
\textbf{\%Broken} &
\textbf{F/B Ratio} &
\textbf{KRR (\%)} \\
\midrule
NRAG & 808 & 50  & 11030 & 12.1 & 1520 & 1.7 & 7.3$\times$ & 97.05 \\
VO   & 30  & 1   & 10993 & 12.2 & 1620 & 1.8 & 6.8$\times$ & 97.79 \\
PAT  & 182 & 3.5 & 11209 & 12.3 & 1518 & 1.6 & 7.4$\times$ & 97.95 \\
\textbf{LAY} &
\textbf{182} & \textbf{3.5} &
\textbf{11238} & \textbf{12.4} &
\textbf{1422$^\ast$} & \textbf{1.6$^\ast$} &
\textbf{7.9$\times$} & \textbf{98.06} \\
\bottomrule
\end{tabular}
\caption{Cluster-level ground per approach. $^\ast$LAY introduces no new breaks; reported breaks persist from earlier grounding stages.}
\label{tab:knowledge_stability_cluster}
\end{table*}

\paragraph{Intervention effectiveness is pattern-dependent.}
Grounding effects vary systematically by structure (Table~\ref{table_hotspot_lift}). CYCLE patterns are the most correctable (0.55 $\rightarrow$ 0.97 under LAY), indicating that structural context resolves directional inconsistencies not captured by verdict grounding alone. TREE patterns are largely resolved through explicit grounding (0.55 $\rightarrow$ 0.93), suggesting that hierarchical reasoning depends primarily on recovering local semantic constraints. STAR patterns remain the most resistant (final accuracy 0.87), reflecting the higher ambiguity induced by hub-spoke semantic organization.
Overall, VO accounts for most improvements, while layered structural grounding provides targeted gains where local dependencies remain unresolved.

This ranking is robust to dataset and grounding condition: on an independent repeated-run evaluation (three domains, NRAG grounding), mean balanced-accuracy improvement is again highest for CYCLE (+0.52) and TREE (+0.27) hotspots and lowest for STAR (+0.09), confirming that pattern type predicts not only where failures concentrate but how much grounding can repair them.

\begin{table*}[t]
\centering
\small
\setlength{\tabcolsep}{6pt}
\begin{tabular}{l r r r r r r}
\hline
\textbf{Pattern} &
\textbf{Baseline} &
\textbf{VO} &
\textbf{LAY} &
\textbf{VO lift} &
\textbf{LAY lift} &
\textbf{LAY -- VO lift} \\
\hline
\textbf{CYCLE} & 0.55 & 0.94 & 0.97 & 0.39 & 0.42 & \textbf{0.03} \\
\textbf{STAR}  & 0.51 & 0.85 & 0.87 & 0.34 & 0.36 & \textbf{0.02} \\
\textbf{TREE}  & 0.55 & 0.93 & 0.93 & 0.38 & 0.38 & \textbf{0.00} \\
\hline
\end{tabular}
\caption{Baseline accuracy and lift comparisons across structural pattern types.}
\label{table_hotspot_lift}
\end{table*}

\paragraph{In summary,}
LLM inconsistencies are structured, sparse, and predictive rather than random. KG grounding improves aggregate performance, but reliable gains require selective, layered intervention guided by hotspot topology and monotonicity constraints.

\section{Related Work}
\label{sec:related_work}

A growing body of work evaluates LLM knowledge and reasoning over large domains~\citep{Jiang2020,Wang2023,yin2023,xie2025,chen2023}, often revealing substantial variability across prompts, runs, and tasks. In particular, prior work on paraphrase consistency and semantic equivalence~\citep{Elazar2021,Uceda_2024,Uceda_2025,Paraphrases,zubiaga2024} shows that LLMs frequently produce inconsistent answers to semantically equivalent queries. Moreover, automated consistency metrics based on resampled outputs (e.g., BERTScore, BLEU, ROUGE) do not necessarily correlate with human judgments~\citep{wu-etal-2025-estimating}, motivating structured, human-curated semantic evaluation.

Recent work has also proposed statistical frameworks for evaluating consistency under semantically preserving perturbations~\citep{Consistency_stat}. These approaches measure reliability through output similarity and trajectory-level stability, showing that high task accuracy may mask substantial behavioral instability. Our work is complementary but differs in both objective and mechanism. Rather than modeling consistency statistically over perturbation distributions, we formalize consistency as semantic invariance over KG-derived equivalence classes. This enables explicit structural localization of failures through hotspot subgraphs and supports selective graph grounding based on ontology topology rather than perturbation sampling alone.

More fundamentally, we establish theoretical limits on semantic invariance: consistency under arbitrary meaning-preserving transformations is undecidable in general. This motivates a shift from global guarantees to \emph{scoped} semantic invariance over bounded semantic domains. Within this setting, we use KGs not only as sources of factual knowledge, but as explicit semantic control structures that define equivalence classes, localize inconsistencies, and expose the structural constraints governing KG-consistent reasoning.

Knowledge graphs have long been used for ontology-based inference and structured reasoning~\citep{Wikidata1,Wikidata2,Wikidata3}. More recently, retrieval-augmented generation (RAG)~\citep{lewis2021,gao2024,edge2025rag} and GraphRAG approaches integrate external knowledge into LLMs for factual grounding and multi-hop reasoning. However, these methods are primarily designed to improve answer completeness or factual accuracy rather than evaluate consistency under semantic equivalence.

Our approach differs in both objective and mechanism. Rather than retrieving broad contextual information, we use KGs to define equivalence classes of queries, identify localized failure regions (hotspot subgraphs), and selectively inject structure-aware grounding. Retrieval is therefore used not to supply missing knowledge, but to expose semantic constraints that the model applies inconsistently. Furthermore, we show that intervention effectiveness is predictable from baseline hotspot structure, a finding not captured in prior RAG or GraphRAG work.

Our work is also related to recent efforts on prompt engineering and LLM orchestration~\citep{LLMon_1}. These approaches improve task performance but typically mix instructions, data, and control flow within unstructured natural-language prompts, making behavior difficult to reproduce and audit across settings. In contrast, we study graph-structured semantic constraints that support localized evaluation and targeted grounding.

Finally, our hotspot decomposition relates to structural motif analysis in graphs~\citep{milo2002network}. However, rather than identifying motifs for network characterization, we use TREE, STAR, and CYCLE structures as semantic signatures of localized LLM inconsistency regions.


\section{Conclusions and Future Work}

We present a framework for evaluating LLM reliability over KG reasoning tasks grounded in a central insight: semantic invariance cannot be guaranteed in general, since consistency under meaning-preserving transformations is undecidable. This motivates a shift from global guarantees to \emph{scoped} semantic invariance over bounded domains.

Within this framework, we show that LLM inconsistencies tend to be structured rather than random and concentrate in TREE, STAR, and CYCLE patterns that remain stable across models and domains (Sections~\ref{sec:scoped_semantic_invariance}, ~\ref{sec:experimental_setup}, ~\ref{sec:results}).

We further show that grounding must be \emph{selective}. While KG augmentation improves aggregate performance, indiscriminate grounding introduces regressions. A layered strategy that applies structural grounding only where needed achieves the strongest monotonicity profile, preserving KG-consistent knowledge while reducing inconsistency.

More broadly, our results suggest a different role for knowledge graphs in LLM systems. Rather than serving only as repositories of external facts, KGs can function as \emph{semantic control structures}: explicit, inspectable interfaces for evaluating, monitoring, and selectively constraining model behavior over bounded domains. In this view, hotspot subgraphs become operational units for semantic auditing, prediction, and targeted grounding.

\paragraph{Broader Implications for NLP.} These results suggest three broader implications. First, aggregate accuracy alone is an insufficient measure of reliability: failures are structured and localized, so evaluation should report where errors occur in a model's semantic space, not only how many occur — a perspective that naturally extends to robustness, factuality, and fairness evaluation. Second, our work illustrates the broader value of explicit semantic representations as an interface between neural models and domain knowledge; while we use knowledge graphs, the underlying principle — grounding model behavior against an explicit semantic specification — is relevant to a wide range of NLP systems, including information extraction, retrieval, dialogue, and neuro-symbolic pipelines. Third, our results suggest a shift in emphasis from maximizing average accuracy to predicting, localizing, and selectively repairing failures; as NLP systems are increasingly deployed in high-stakes settings, methods that identify where models are unreliable and apply targeted interventions may become as important as improving aggregate benchmark performance.

\paragraph{Future Work.}
A natural extension is the development of lightweight KG-grounded monitoring and explanation systems. Because hotspot subgraphs are sparse, compact, and structurally regular, we could index and fetch them efficiently without heavy semantic retrieval infrastructure. More generally, structured semantic interfaces may provide a foundation for more disciplined and reproducible interaction with LLMs in high-stakes and domain-specific settings.                                       
As a step in this direction, we are developing an extended benchmark that moves beyond yes/no queries toward more realistic multi-hop questions that require explicit reasoning over KG structure, such as property inheritance along subconcept chains, directionality disambiguation in near-cycles, and hub-node enumeration. A preview of this benchmark, including the evaluation results reported in this paper, is provided in Appendix~\ref{appendix:testing_realistic_queries}.

\section*{Limitations}

Our framework evaluates semantic consistency relative to explicit KG structure, and therefore inherits the assumptions and limitations of the underlying ontology. Inconsistencies identified by the framework do not necessarily imply that the model is wrong: in some cases, the KG itself may encode debatable, incomplete, or counterintuitive relations, particularly in open or collaboratively curated sources such as Wikidata. The framework should therefore be interpreted as measuring alignment with a controlled semantic specification rather than absolute correctness.

This work focuses primarily on ontology-centered reasoning, especially hierarchical and relational consistency over \texttt{IsA}-style structures. Although the framework extends naturally to richer relations, our experiments do not yet evaluate broader forms of reasoning such as temporal, causal, procedural, or multi-hop explanatory inference.

Our intervention results should likewise be interpreted cautiously. Verdict-only (VO) grounding assumes direct access to the queried edge and therefore functions as an oracle condition rather than a deployable retrieval strategy. While the layered approach reduces regressions and improves monotonicity, practical deployment still depends on reliably mapping user queries to relevant KG regions.

Finally, hotspot patterns capture localized structural failures, but do not fully explain the internal mechanisms producing those failures. The observed regularities show that inconsistency is constrained by KG topology, yet the representational causes inside the models remain an open question.

\bibliography{reliability_of_LLMs_ref}

\appendix

\section{Undecidability of Semantic Invariance}
\label{appendix:undecidability}

This appendix formalizes the theoretical claim underlying the main paper: semantic invariance cannot, in general, be guaranteed for open-domain LLMs.

\subsection{Formal Setup}

Let $X$ denote a set of well-formed natural language queries and let $A=\{\text{yes},\text{no}\}$ be a finite answer space. We model an LLM as a (possibly randomized) function
\[
F : X \times \Omega \rightarrow A,
\]
where $\omega \in \Omega$ captures stochasticity in decoding.

We assume a semantic equivalence relation $\sim$ over $X$, where
\[
q_1 \sim q_2
\]
means that $q_1$ and $q_2$ express the same underlying meaning.

\begin{definition}[Semantic Invariance]
A model $F$ is semantically invariant with respect to $\sim$ if $\forall q_1,q_2 \in X$
\[
\begin{aligned}
q_1 \sim q_2 \;\Rightarrow\;& \forall \omega_1,\omega_2 \in \Omega, \\
                           & F(q_1,\omega_1)=F(q_2,\omega_2)
\end{aligned}
\]
\end{definition}

Intuitively, semantically equivalent inputs should always produce identical outputs, regardless of decoding stochasticity.

\subsection{Undecidability of General Semantic Invariance}

Guaranteeing semantic invariance requires deciding whether two queries induce the same semantic behavior under the model. For unrestricted semantic domains, this becomes a non-trivial equivalence problem.

Consider a Turing-complete programming language. For each program $P_i$, define a corresponding natural-language query:
\begin{center}
\emph{``Does $P_i$ halt on all inputs?''}
\end{center}

Each such query has a well-defined semantic truth value. Define semantic equivalence between queries by:
\[
q_{P_1} \sim q_{P_2}
\]
iff $P_1$ and $P_2$ share the same halting behavior.

Suppose there existed a general procedure capable of deciding semantic equivalence for arbitrary queries in $X$. Then, given programs $P_1$ and $P_2$, we could determine whether they share the same semantic property with respect to halting behavior.

However, by Rice’s theorem~\citep{FoundationsComputabilityTheory,Rice_1}, every non-trivial semantic property of sufficiently expressive programs is undecidable. Consequently, determining semantic equivalence over unrestricted query spaces is undecidable as well.

This implies that no general algorithmic procedure can guarantee semantic invariance for open-domain LLMs operating over unrestricted natural language.

\subsection{Restricted Formal Languages}

The limitation persists even under strong restrictions. Query equivalence remains undecidable for expressive fragments of relational query languages such as Select-Project-Join-Union-Difference (SPJUD)~\citep{SQL_0,SQL_1,SQL_2}, despite being substantially simpler than unrestricted natural language.

Thus, the source of the limitation is not merely linguistic ambiguity, but the expressive power of the underlying semantic domain.

\subsection{Implications for LLMs}

The undecidability result is independent of stochastic decoding or incomplete model knowledge. Even under deterministic decoding (e.g., temperature $=0$), enforcing semantic invariance over unrestricted semantic domains remains impossible in general.

These observations motivate the scoped formulation adopted in the main paper. Rather than attempting to guarantee invariance over unrestricted natural language, we evaluate consistency relative to finite equivalence classes induced by explicit semantic structure in a knowledge graph.


\section{Dataset Construction and Hotspot Extraction}
\label{appendix:dataset_construction}

This appendix describes the construction of \textsc{SI-Cyber}, the extraction of hotspot subgraphs, and the structural organization of failures observed across the D3FEND ontology.

\subsection{SI-Cyber Dataset Construction} 
\label{sec:SI-Cyber}

We construct \textsc{SI-Cyber}, a dataset derived from the MITRE D3FEND ontology~\citep{mitre_d3fend_ontology}, a real-world cybersecurity KG used for defensive reasoning and operational security analysis. The D3FEND ontology integrates with governance frameworks such as NIST SP~800-53 and DISA Control Correlation Identifiers (CCIs), making it suitable for evaluating semantic consistency over mission-critical cybersecurity knowledge. This dataset is available at \url{https://huggingface.co/datasets/ibm-research/SI-Cyber}.

Our construction follows the KG-based methodology introduced for Wikidata-derived datasets in~\cite{Uceda_2024,Uceda_2025}, extending it from partially synthesized domains to a substantially richer operational ontology.

The dataset is organized into \emph{query clusters}, where each cluster probes an ontological relation through multiple paraphrased formulations. This enables systematic evaluation of semantic consistency across logically equivalent queries. The resulting cluster types span a progression from direct edge lookup to hierarchical and relational reasoning.

\paragraph{Root Selection.}
We first select a set of productive root concepts:

\begin{itemize}
\item d3f:Resource
\item d3f:File
\item d3f:Process
\item d3f:NetworkNode
\item d3f:Software
\item d3f:Credential
\item d3f:DefensiveTechnique
\end{itemize}

We then select ontology relations suitable for inheritance-based reasoning. In our experiments, we focus on \texttt{d3f:connected-to} and \texttt{d3f:contains}, both of which induce meaningful relational constraints over subclasses.

\subsection{Hierarchy Extraction}

The SPARQL query in Figure~\ref{fig:d3fend_fanout_query} extracts \texttt{rdfs:subClassOf} relations rooted at the selected concepts. Since D3FEND contains roughly 34,500 tuples (excluding deductive closure), we additionally filter for usable labels, non-deprecated nodes, and non-trivial fan-out or descendant structure.

This extraction supports four cluster types:

\begin{enumerate}
    \item \texttt{edge-yes}: direct ontology edges,
    \item \texttt{edge-no}: verified non-edges,
    \item \texttt{edge-inv}: inverse-edge asymmetry tests,
    \item \texttt{hierarchy}: transitive inheritance paths.
\end{enumerate}

For inverse-edge clusters, we additionally verify that no symmetric or anti-symmetric ontology property invalidates the construction.

\begin{figure}[!t]
\centering
\begin{minipage}{0.95\columnwidth}
\begin{lstlisting}[
  basicstyle=\ttfamily\small,
  breaklines=true,
  columns=fullflexible,
  keepspaces=true
]
PREFIX rdfs: <http://www.w3.org/2000/01/rdf-schema#>
PREFIX d3f:  <http://d3fend.mitre.org/ontologies/d3fend.owl#>

SELECT ?root ?rootLabel ?parent ?parentLabel ?child ?childLabel
       (COUNT(DISTINCT ?grandchild) AS ?childFanout)
       (COUNT(DISTINCT ?desc) AS ?descendantCount)
WHERE {
  VALUES ?root {
    d3f:Resource
    d3f:File
    d3f:Process
    d3f:NetworkNode
    d3f:Software
    d3f:Credential
    d3f:DefensiveTechnique
  }

  ?child rdfs:subClassOf ?parent .
  ?parent rdfs:subClassOf* ?root .

  OPTIONAL { ?root   rdfs:label ?rootLabel }
  OPTIONAL { ?parent rdfs:label ?parentLabel }
  OPTIONAL { ?child  rdfs:label ?childLabel }
  OPTIONAL { ?grandchild rdfs:subClassOf ?child . }

  OPTIONAL {
    ?desc rdfs:subClassOf+ ?child .
    FILTER(?desc != ?child)
  }
}
GROUP BY ?root ?rootLabel ?parent ?parentLabel ?child ?childLabel
HAVING (COUNT(DISTINCT ?grandchild) > 0 ||
         COUNT(DISTINCT ?desc) > 0)
ORDER BY ?root DESC(?descendantCount)
         DESC(?childFanout)
         ?parentLabel ?childLabel
\end{lstlisting}
\end{minipage}
\caption{SPARQL query computing fan-out and descendant counts in the D3FEND ontology.}
\label{fig:d3fend_fanout_query}
\end{figure}

\subsection{Property-Hierarchy Extraction}

The SPARQL query in Figure~\ref{fig:sparql_query_relationships} generates \texttt{property-hierarchy} clusters, which test whether properties are inherited through subclass relations. Given:

\[
\begin{aligned}
(A,\texttt{subconceptOf},B), \\
(B,\texttt{subconceptOf},C), \\
(C,\texttt{propertyX},D)
\end{aligned}
\]
the model should infer:

\[
(A,\texttt{propertyX},D).
\]

We focus on relations likely to induce inheritance behavior in LLMs, such as \texttt{contains} and \texttt{connected-to}. Figure~\ref{fig:sparql_query_relationships} shows the extraction query for \texttt{connected-to}; the \texttt{contains} query substitutes the corresponding relation.

\begin{figure}[!t]
\centering
\begin{minipage}{0.95\columnwidth}
\begin{lstlisting}[
  basicstyle=\ttfamily\small,
  breaklines=true,
  columns=fullflexible,
  keepspaces=true
]
PREFIX rdf:  <http://www.w3.org/1999/02/22-rdf-syntax-ns#>
PREFIX rdfs: <http://www.w3.org/2000/01/rdf-schema#>
PREFIX owl:  <http://www.w3.org/2002/07/owl#>
PREFIX d3f:  <http://d3fend.mitre.org/ontologies/d3fend.owl#>

SELECT DISTINCT ?a ?aLabel ?b ?bLabel ?c ?cLabel
WHERE {
  ?a rdfs:subClassOf+ ?b .
  FILTER(?a != ?b)

  {
    ?b d3f:connected-to ?c .
  }
  UNION {
    ?b rdfs:subClassOf ?restr .
    ?restr a owl:Restriction ;
           owl:onProperty d3f:contains ;
           owl:someValuesFrom ?c .
  }

  OPTIONAL { ?a rdfs:label ?aLabel }
  OPTIONAL { ?b rdfs:label ?bLabel }
  OPTIONAL { ?c rdfs:label ?cLabel }

  FILTER(isIRI(?a) && isIRI(?b) && isIRI(?c))
}
ORDER BY ?bLabel ?aLabel ?cLabel
\end{lstlisting}
\end{minipage}
\caption{SPARQL query used to retrieve relational constraints in the D3FEND ontology.}
\label{fig:sparql_query_relationships}
\end{figure}

The resulting dataset composition is summarized in Table~\ref{tab:cyber_dataset_decomposition} and Table~\ref{tab:query_counts} (\textsc{SI-Cyber} row only).

\begin{table*}[!ht]
    \centering
    \setlength{\tabcolsep}{4pt}
    \begin{tabular}{lrp{7.5cm}}
    \toprule
    \textbf{Cluster type} & \textbf{Clusters} & \textbf{Description} \\
    \midrule
    \texttt{edge-yes}           & 2,045 & Direct subconcept edge in the ontology.
                                           Expected: \textit{yes}. \\
    \texttt{edge-inv}           & 2,045 & Inverse of a direct edge; tests asymmetry
                                           of the subconcept relation.
                                           Expected: \textit{no}. \\
    \texttt{edge-no}            & 1,227 & Verified non-edge between co-occurring concepts.
                                           Expected: \textit{no}. \\
    \texttt{hierarchy}          &   720 & Transitive (multi-hop) subconcept relationship;
                                           requires reasoning over intermediate nodes.
                                           Expected: \textit{yes}. \\
    \texttt{property-hierarchy} & 1,943 & Relationship via a non-taxonomic property
                                           (e.g., \texttt{connected\_to}); requires
                                           relational semantics beyond pure taxonomy.
                                           Expected: \textit{yes}. \\
    \midrule
    \textbf{Total}              & 7,980 & \\
    \bottomrule
    \end{tabular}
    \caption{Dataset composition for the \textsc{SI-cyber} domain.}
    \label{tab:cyber_dataset_decomposition}
  \end{table*}   

\subsection{Hotspot Extraction in SI-Cyber}
\label{appendix:cyber-hotspots}

The hotspot analysis of \textsc{SI-Cyber} identifies 200 hotspot edges grouped into 24 connected hotspot subgraphs. Every hotspot edge is flagged by at least 5 of the 6 evaluated models (hot fraction $\geq 5/6$), and 110 edges (55\%) are flagged by all six models, indicating substantially stronger cross-model agreement than in the Wikidata domains.

three dominant hotspot regions contain most failures, while the remaining 21 subgraphs are small (median size: 2 edges). Table~\ref{tab:cyber_hs_subgraphs} summarizes the five largest hotspot regions.

\begin{table*}[!ht]
\centering
\begin{tabular}{lrrrr}
\toprule
\textbf{Subgraph} & \textbf{Nodes} & \textbf{Edges} & \textbf{Hot Frac.} & \textbf{Consistency} \\
\midrule
HS\_0007 & 25 & 50 & 1.000 & 0.757 \\
HS\_0003 & 39 & 40 & 1.000 & 0.707 \\
HS\_0006 & 27 & 26 & 1.000 & 0.741 \\
HS\_0004 & 22 & 23 & 1.000 & 0.688 \\
HS\_0010 & 13 & 13 & 1.000 & 0.719 \\
\bottomrule
\end{tabular}
\caption{Top-5 hotspot subgraphs in \textsc{SI-cyber}, ranked by edge count.
Mean balanced accuracy is not available at the subgraph level since per-edge
balanced accuracy requires per-model data (see Table~\ref{table_per_model}).}
\label{tab:cyber_hs_subgraphs}
\end{table*}

\subsection{Concentrated Failure Axes}

Hotspot targets concentrate around two major semantic regions shown in Table~\ref{tab:target_concentration}: taxonomic (subConceptOf) and relational (connected-to) failures. 

\begin{table*}[!ht]
\centering
\begin{tabular}{lrrl}
\toprule
\textbf{Target concept} & \textbf{\# Edges} & \textbf{Property} & \textbf{Failure axis} \\
\midrule
Defensive Technique & 30 & subConceptOf & Axis 1 — Taxonomic \\
Executable File     & 26 & connected to & Axis 2 — Relational \\
Software            & 23 & subConceptOf & Axis 1 — Taxonomic \\
Firmware            & 18 & connected to & Axis 2 — Relational \\
Operating System    & 15 & connected to & Axis 2 — Relational \\
Hardware Device     & 15 & connected to & Axis 2 — Relational \\
Network Node        & 11 & subConceptOf & Axis 1 — Taxonomic \\
File Section        &  9 & connected to & Axis 2 — Relational \\
Resource            &  5 & subConceptOf & Axis 1 — Taxonomic \\
Remote Resource     &  3 & subConceptOf & Axis 1 — Taxonomic \\
\midrule
\textit{Axis subtotal} & 155 & \multicolumn{2}{l}{\textit{(78\% of 200 edges)}} \\
\bottomrule
\end{tabular}
\caption{Top-10 target nodes in hotspot edges for SI-CYBER, grouped by failure axis. The two axes together account for 78\% of all hotspot edges.}
\label{tab:target_concentration}
\end{table*}

\paragraph{Axis 1 — Taxonomic failures.}
Models systematically fail on subConceptOf/hierarchy edges — subtype and transitive-hierarchy relations. This spans digital-artifact classes (Software, Network Node, Resource, Remote Resource) and defensive-technique classes such as Disk Partitioning and Relay Pattern Analysis, which are frequently not recognized as subconcepts of Defensive Technique despite explicit ontology membership.

\paragraph{Axis 2 — Relational (connected-to) failures}
Models systematically fail on connected to edges linking concrete devices, firmware, and executables to their structural parents — e.g., routers, servers, and OT controllers disconnected from firmware/OS/hardware parent classes, and low-level API/syscall constructs disconnected from Executable File. The largest hotspot region (HS\_0007) spans 50 edges and 25 nodes. For example, every model fails Firewall $\rightarrow$ Hardware Device, Firewall $\rightarrow$ Operating System, and Firewall $\rightarrow$ Firmware, despite these relations being explicitly represented in the ontology.

A striking pattern is that all six models achieve extremely high specificity (0.960–0.996), correctly rejecting false edges while failing primarily through omission rather than hallucination. Errors are therefore strongly asymmetric: models are conservative and frequently fail to assert true ontology relations.

\section{Prompting and Evaluation Protocol}
\label{appendix:prompting_protocol}

This appendix describes prompting configurations, model-specific settings, evaluation metrics, and intervention protocols.

\subsection{Grounding Configurations}

We evaluate the following prompting conditions:

\paragraph{BL (Baseline).}
Queries are presented without additional grounding context.

\paragraph{VO (Verdict-Only).}
The target KG relation is injected directly as an explicit semantic constraint.

\paragraph{PAT (Pattern-Aware).}
VO is augmented with local structural context derived from hotspot patterns (TREE, STAR, or CYCLE).

\paragraph{NRAG (Naive RAG).}
A naive RAG approach where where the document set consists of the manually curated natural language (English) comments associated with each edge in the hotspot pattern. These comments are part of the metadata that already exist within the D3FEND and Wikidata ontologies.

\paragraph{LAY (Layered).}
Grounding is applied selectively: stronger structural context is introduced only when simpler grounding fails.

Results are reported for single run for each condition.

\subsection{Prompt Templates}

\paragraph{Baseline Prompt.}

\begin{quote}\ttfamily
You are a strict yes/no classifier.\\
Respond with exactly one word: Yes or No.\\[4pt]
Question: \{question\}\\
Answer:
\end{quote}

\paragraph{VO/PAT Shared Prompt Frame.}

\begin{quote}\ttfamily
You are a knowledge graph verifier. Your only job is to report what the knowledge graph says.\\[4pt]

RULES:\\
1. The knowledge graph is the sole source of truth.\\
2. Answer with exactly one word: Yes or No.\\
3. A relationship A is-a B means A is a subconcept/subtype of B.\\
4. Watch for directionality.\\
5. If a DIRECT KG VERDICT is shown, it is authoritative.\\[4pt]

\{evidence\}\\[4pt]

Question: \{question\}\\
Answer:
\end{quote}

\paragraph{VO Evidence Block.}

\begin{quote}
\texttt{DIRECT KG VERDICT:}\\
\texttt{[YES] (Container Build Tool) --[subclass of]--> (Software Packaging Tool)}
\end{quote}

\paragraph{PAT Evidence Block.}

\begin{quote}
\texttt{DIRECT KG VERDICT:}\\
\texttt{[YES] (Container Build Tool) --[subclass of]--> (Software Packaging Tool)}\\[4pt]
\end{quote}

\subsection{Model-Specific Settings}

All experiments use temperature $=0$.
Each of the six models required minor prompt engineering to elicit reliable yes/no answers. For most models, appending an explicit answer trigger such as "Answer:" or "Answer (Yes or No):" at the end of the prompt was sufficient to suppress verbose or hedged responses and force a single-token reply, keeping max tokens at 20. DeepSeek-V3.2 required a more directive trigger phrasing to reduce refusals on binary questions. Qwen3.5-397B needed thinking mode explicitly disabled via enable-thinking: false in the chat template to avoid chain-of-thought preamble before the answer. GPT-OSS-120B was the exception: it reliably produced short answers without a trigger, but required the Harmony chat template and a higher token budget of 500, reflecting its different instruction-following behavior under that serving configuration. Table~\ref{tab:model_settings} summarizes the variations on the prompting strategies, which were adopted on a trial-and-error basis. 

  \begin{table*}[!ht]
    \centering
    
    \setlength{\tabcolsep}{4pt}
    \begin{tabular}{lcclll}
    \toprule
    \textbf{Model} & \textbf{\# Params} & \textbf{Max tokens} & \textbf{Answer trigger} & \textbf{Chat template} \\
    \midrule
    \href{https://huggingface.co/openai/gpt-oss-120b}{gpt-oss-120b} & 120B & 500          & (none) & harmony   \\
    \href{https://huggingface.co/meta-llama/Llama-3.3-70B-Instruct}{Llama-3.3-70B} & 70B  & 20       & \texttt{Answer:}     & default     \\
    \href{https://huggingface.co/Qwen/Qwen3.5-397B-A17B-FP8}{Qwen3.5-397B} & 397B   & 20       & (none) & default*  \\
    \href{https://huggingface.co/deepseek-ai/DeepSeek-V3.2}{DeepSeek-V3.2} & 685B  & 20      & \texttt{Answer (Yes or No):} & deepseek \\
    \href{https://huggingface.co/ibm-granite/granite-4.0-h-small}{Granite-4.0-h-small} & 32B & 20 & \texttt{Answer:}     & default   \\
    \href{https://huggingface.co/mistralai/Mistral-Small-3.2-24B-Instruct-2506}{Mistral-Small-24B} & 24B & 20    & \texttt{Answer:}     & mistral    \\
    \bottomrule
    \end{tabular}
    \caption{Models, model parameters, and model-specific prompt settings (max output tokens, answer trigger, and chat template). *For Qwen3.5-397B, the default chat template was used with \texttt{enable\_thinking} set to \texttt{false}.}
     \label{tab:model_settings}
  \end{table*}

\subsection{Query-Level Grounding Analysis}
\begin{table*}[!t]
\centering
\small
\setlength{\tabcolsep}{4pt}
\begin{tabular}{lrrrrrrrr}
\toprule
\textbf{Approach} &
\textbf{Avg Tokens} &
\textbf{Avg Triples} &
\textbf{Fixed} &
\textbf{\%Fixed} &
\textbf{Broken} &
\textbf{\%Broken} &
\textbf{F/B Ratio} &
\textbf{KRR (\%)} \\
\midrule
NRAG & 808 & 50.00 & 61768 & 12.80 & 8367 & 1.80 & 7.26$\times$ & 98.20 \\
VO   & 30  & 1.00  & 69276 & 14.40 & 6973 & 1.50 & 9.93$\times$ & 98.40 \\
PAT  & 182 & 3.50  & 70443 & 14.50 & 6981 & 1.40 & 10.09$\times$ & 98.50 \\
\textbf{LAY} &
\textbf{182} & \textbf{3.50} &
\textbf{70502} & \textbf{14.70} &
\textbf{6149$^\ast$} & \textbf{1.30$^\ast$} &
\textbf{11.47$\times$} & \textbf{98.60} \\
\bottomrule
\end{tabular}
\caption{Query-level grounding per approach. $^\ast$LAY introduces no new breaks; reported breaks persist from earlier grounding stages.}
\label{tab:knowledge_stability_query}
\end{table*}
Query-level results (in~\ref{tab:knowledge_stability_query}) are consistently more favorable than cluster-level ones because grounding tends to repair multiple queries within failing clusters while breaking only isolated queries within otherwise KG-consistent regions.

\subsection{Deployment Caveat}

LAY assumes oracle knowledge of which clusters BL already answers consistently. In practical systems, this decision must be approximated through retrieval or learned routing. In this sense, hotspot-aware grounding replaces global prompting with selective structural intervention.

\subsection{Models Sizes \& Budget}

Model sizes are reported in Table \ref{tab:model_settings}. Because we test models of varying sizes and from different families, we report the total number of queries (i.e. inference calls) across all datasets and the estimated number of input tokens per inference call (Table \ref{tab:query_counts}). The \textsc{SI-Cyber} dataset contains 7980 clusters (as also noted in Section \ref{sec:experimental_setup}), with 5-6 questions per cluster, this results in roughly 45k queries and therefore 45k inferences calls to each model. For the 8 Wikidata domains (academic disciplines, dishes, finance, natural disasters, home appliances, medical specialties, software, and music genres) derived from the datasets introduced in \cite{Uceda_2024, Uceda_2025}, each model was queried 35,560 times. Altogether, combined with the 25 manually curated complex queries, the total number of inference calls per model per prompting configuration for all three datasets is 80,817. The figures in Table \ref{tab:query_counts} provide an infrastructure- and model-agnostic sense of the inference requirements of our experiments.

  \begin{table*}[!ht]
    \centering
    
    \setlength{\tabcolsep}{4pt}
    \begin{tabular}{lccc}
    \toprule
    \textbf{Dataset} & \textbf{\# Question Clusters} & \textbf{\# Questions} & \textbf{\# Input Tokens} \\
    \midrule
    \textsc{SI-Cyber} & 7,980 & 45,232 & 519,187 \\
    Wikidata Domains & 7,244 & 35,560 & 367,353 \\
    Curated complex queries & n/a & 25 & 1,540 \\
    \midrule
    \textbf{Total} & 15,224 & 80,817 & 888,080 \\
    \bottomrule
    \end{tabular}
    \caption{Number of question clusters, questions (i.e. inference calls), and input tokens for each of the three datasets used. These provide an infrastructure- and model-agnostic picture of the inference requirements of our experiments.}
     \label{tab:query_counts}
  \end{table*}

\subsection{Sensitivity to Decoding Randomness} \label{sec:sensitivity_to_randomness}
 To assess sensitivity to decoding randomness, we repeated the baseline evaluation for three models (DeepSeek-V3.2, Granite-4.0-h-small, Mistral-Small-3.2-24B-Instruct) across three domains (finance, software, SI-Cyber) with K=3 independent runs at temperature 0.3 (versus temperature 0 in the main results). Balanced accuracy and consistency are stable across repeats (finance: 0.831$\pm$0.001 / 0.963$\pm$0.001; software: 0.820$\pm$0.000 / 0.957$\pm$0.001; SI-Cyber: 0.739$\pm$0.001 / 0.873$\pm$0.000), and the TREE/STAR/CYCLE pattern distribution varies by at most 8 percentage points across runs within a domain, with the dominant pattern type unchanged. Comparing temperature 0 to 0.3 directly on the same clusters, balanced accuracy shifts by at most 0.030 and consistency by at most 0.014 across all nine (model, domain) pairs, confirming our single-run, temperature-0 results are not an artifact of decoding-randomness averaging.

We additionally check whether hotspot edges are shared across the six evaluated models rather than being model-specific artifacts. Restricting to the hotspot set (edges flagged hot by at least two of six models), we find graded but substantial cross-model agreement: across the nine domains, a mean of 46\% of hotspot edges are flagged by at least five of six models, and a mean of 27\% are flagged by all six. Agreement is strongest in Finance, where 67.7\% of hotspot edges are shared by at least five models and 51.1\% by all six models. This indicates that hotspot status reflects a structural property of the underlying KG topology rather than idiosyncratic behavior of any single model, though — consistent with our framing throughout — agreement is graded rather than uniform across domains.

\section{Extended Results and Complex Queries}
\label{appendix:extended_results}

\subsection{Extended Domain and Model Results}

We evaluate six models across nine domains under the four grounding conditions: BL, VO, PAT, and LAY.

\paragraph{Overall gains.}
Averaged across domains, BL achieves 82.4\% balanced accuracy. VO raises this to 91.4\%, PAT to 92.8\%, and LAY to 95.5\%.

\paragraph{Domain variation.}
Performance gains vary substantially by domain. In \texttt{software} and \texttt{medical\_specialties}, VO yields large improvements, indicating limited parametric coverage. In \texttt{academic\_disciplines}, BL already approaches ceiling performance.

The most instructive case is \texttt{natural\_disasters}, where VO regresses below BL, introducing confounding context. LAY fully recovers this regression by selectively preserving KG-aligned BL clusters.


\begin{figure*}[!t]
  \centering
  \includegraphics[width=\linewidth]{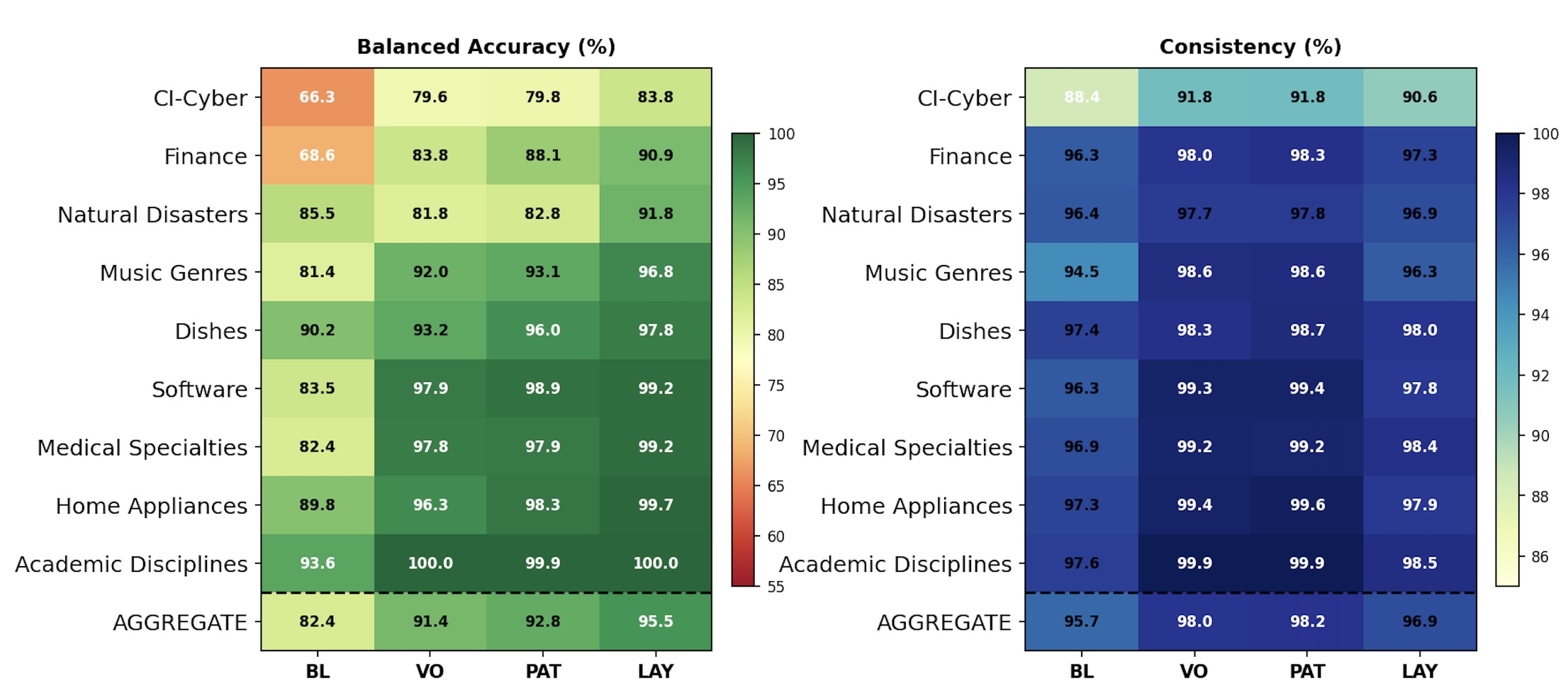}
  \caption{Balanced accuracy and consistency improvements from baseline (BL) for Verdict-Only (VO), Pattern (PAT) and Layered (LAY) averaged across domains.}
  \label{table_per_domain}
\end{figure*}

\paragraph{Model-Level Results}

\begin{figure*}[!t]
  \centering
  \includegraphics[width=\linewidth]{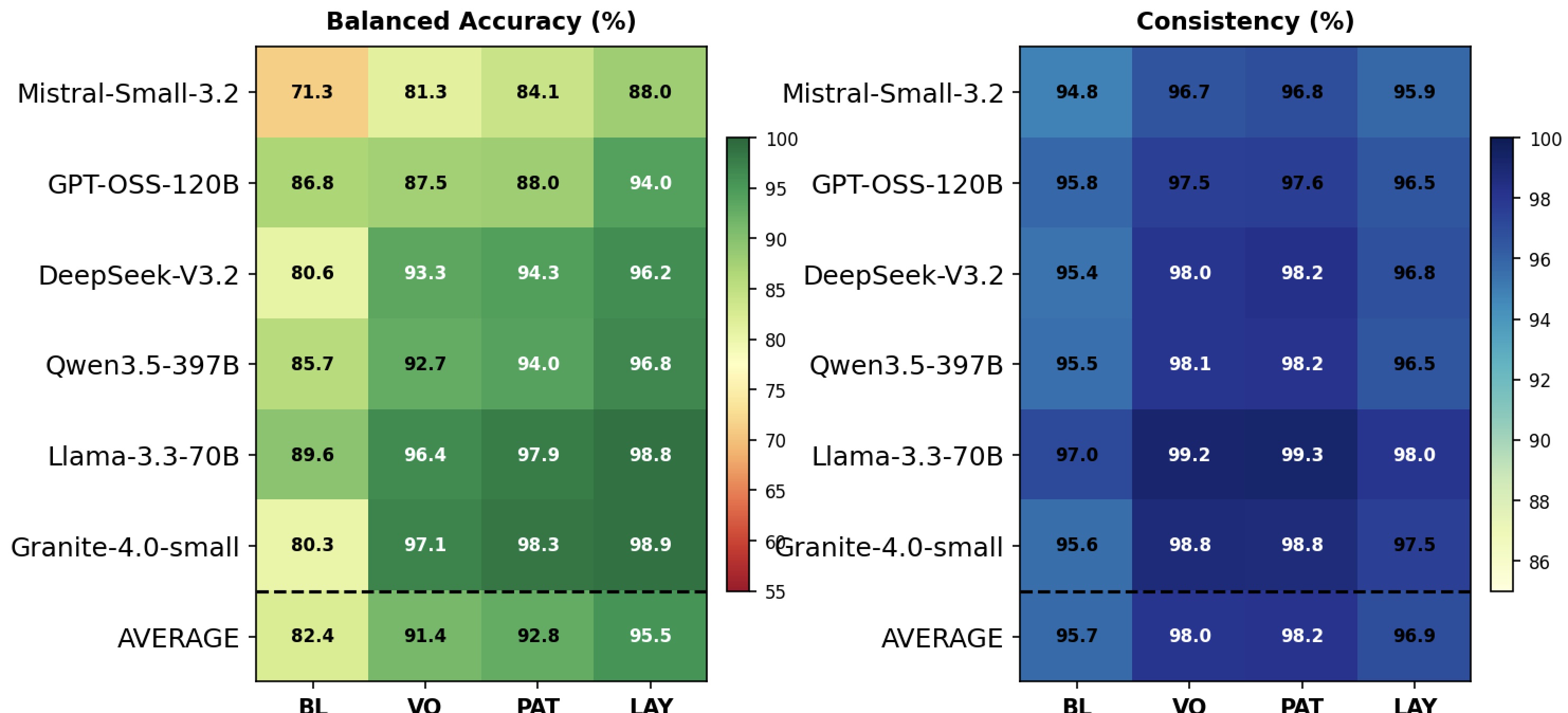}
  \caption{Balanced accuracy and consistency improvements from baseline (BL) for Verdict-Only (VO), Pattern (PAT) and Layered (LAY) per model.}
  \label{table_per_model}
\end{figure*}

LLaMA-3.3-70B and Granite-4.0-h-small achieve the highest LAY performance. Mistral-Small-24B has the weakest BL performance but the largest LAY gain, indicating that weaker models benefit most from layered grounding (as seen in fig~\ref{table_per_model}).

\subsection{Detailed Hotspot Examples}

This section provides detailed examples of the three hotspot structures discussed in the main paper.

\paragraph{CYCLE: Liturgical Music}

The cycle between \textit{church music} and \textit{liturgical music} illustrates conceptual collapse. Models frequently reverse the ontology direction, behaving as if both concepts were equivalent.

\begin{table*}[!t]
\centering
\small
\begin{tabular}{lcccc}
\toprule
\textbf{Edge} & \textbf{Agg.\ acc} & \textbf{Agg.\ cons} & \textbf{n\_hot} & \textbf{Hot?} \\
\midrule
church music $\rightarrow$ liturgical music (exp.: yes) & 0.58 & 0.75 & 5/6 & Yes \\
liturgical music $\rightarrow$ church music (exp.: no)  & 0.08 & 0.92 & 6/6 & Yes \\
\bottomrule
\end{tabular}
\caption{Aggregate accuracy, consistency, hot fraction of Cycle between Liturgical music and Church music}
\label{tab:cycle_detailed_results}
\end{table*}

\paragraph{STAR: Pediatric Neurosurgeon}

The STAR pattern centered on \textit{pediatric neurosurgeon} illustrates failures of semantic generalization across related fields.

\begin{table*}[!t]
\centering
\small
\begin{tabular}{lcccc}
\toprule
\textbf{Edge} & \textbf{Agg.\ acc} & \textbf{Agg.\ cons} & \textbf{n\_hot} & \textbf{Hot?} \\
\midrule
pediatric\_neurosurgeon $\rightarrow$ childhood disease & 0.67 & 0.67 & 5/6 & Yes \\
pediatric\_neurosurgeon $\rightarrow$ neurology         & 0.85 & 0.85 & 3/6 & Yes \\
pediatric\_neurosurgeon $\rightarrow$ pediatric surgery & 0.90 & 0.90 & 2/6 & Yes \\
pediatric\_neurosurgeon $\rightarrow$ pediatrics        & 0.92 & 0.92 & 2/6 & Yes \\
\bottomrule
\end{tabular}
\caption{Aggregate accuracy, consistency, hot fraction of STAR with hub neurosurgeon}
\label{tab:star_detailed_results}
\end{table*}

\paragraph{TREE: GNU/Linux Hierarchy}

The TREE pattern rooted in GNU/Linux concepts illustrates failures of transitive inheritance across software hierarchies.

\begin{table*}[!t]
\centering
\small
\begin{tabular}{lcccc}
\toprule
\textbf{Edge} & \textbf{Agg.\ acc} & \textbf{Agg.\ cons} & \textbf{n\_hot} & \textbf{Hot?} \\
\midrule
GNU Project $\rightarrow$ operating system                & 0.75 & 0.75 & 6/6 & Yes \\
GNU Project $\rightarrow$ Unix-like OS                    & 0.75 & 0.75 & 6/6 & Yes \\
GNU Project $\rightarrow$ GNU/Linux                       & 0.71 & 0.71 & 5/6 & Yes \\
Linux $\rightarrow$ operating system                      & 0.83 & 0.83 & 4/6 & Yes \\
GNU/Linux $\rightarrow$ operating system                  & 0.88 & 0.88 & 3/6 & Yes \\
Linux $\rightarrow$ freely redistributable software       & 0.85 & 0.85 & 2/6 & Yes \\
GNU Project $\rightarrow$ Linux                           & 0.88 & 0.88 & 2/6 & Yes \\
Linux $\rightarrow$ system software                       & 0.88 & 0.88 & 2/6 & Yes \\
Linux $\rightarrow$ computer science                      & 0.85 & 0.85 & 2/6 & Yes \\
GNU Project $\rightarrow$ system software                 & 0.84 & 0.84 & 2/6 & Yes \\
Linux $\rightarrow$ free software                         & 0.92 & 0.92 & 2/6 & Yes \\
Linux $\rightarrow$ source-available software             & 0.92 & 0.92 & 2/6 & Yes \\
\bottomrule
\end{tabular}
\caption{Aggregate accuracy, consistency, hot fraction of TREE from Operating System}
\label{tab:tree_detailed_results}
\end{table*}

\subsection{Complex Graph-Grounded Queries}
\label{appendix:testing_realistic_queries}

The experiments above evaluate semantic invariance using controlled yes/no query clusters derived directly from KG structure. To test whether the same hotspot patterns transfer to more realistic reasoning settings, we constructed a small set of human-authored compositional queries grounded in the same Wikidata domains and semantic structures.

We evaluate five LLMs on 25 complex queries drawn from six domains used in Section~\ref{sec:experimental_setup} (finance, medical specialties, academic disciplines, dishes, music genres, and software). This is a small, preliminary dataset (N=25) intended as a targeted confirmatory check — testing whether TREE, STAR, and CYCLE patterns identified via controlled equivalence-cluster probes also appear in realistic, human-authored compositional queries — rather than a large-scale validation of the hotspot taxonomy; we leave a larger-scale complex-query evaluation to future work. Unlike the controlled cluster evaluation, these questions embed the target relation within short natural-language scenarios requiring conceptual or hierarchical reasoning. Examples include:

\begin{quote}
``The insurance form only covers consultations with surgeons. Would an appointment with a pediatric neurosurgeon be eligible?''
\end{quote}

\begin{quote}
``How do I file my statements for my health savings account, under insurance or under bonds?''
\end{quote}

The query types map directly to hotspot structures:

\noindent\textbf{Patterns:}
(1) \textbf{Transitive justification} (TREE),
(2) \textbf{Generality with evidence} (TREE),
(3) \textbf{Cycle directionality} (CYCLE),
(4) \textbf{Graph explanation hub} (STAR).

\begin{table*}[!t]
\centering
\small
\begin{tabular}{lcccc}
\toprule
\textbf{Metric} & \textbf{BL} & \textbf{VO} & \textbf{PAT} & \textbf{LAY} \\
\midrule
Raw Accuracy            & 0.690 & 0.920 & 0.950 & 0.960 \\
Balanced Accuracy       & 0.515 & 0.900 & 0.903 & 0.919 \\
Cross-model Consistency & 0.868 & 0.948 & 0.974 & --    \\
\bottomrule
\end{tabular}
\caption{Ensemble (majority vote across five models).}
\label{tab:ensemble_results}
\end{table*}

\begin{table*}[!t]
\centering
\small
\begin{tabular}{lcccc}
\toprule
\textbf{Type} & \textbf{BL} & \textbf{VO} & \textbf{PAT} & \textbf{LAY} \\
\midrule
cycle\_directionality     & 0.750 & 1.000 & 0.938 & 1.000 \\
generality\_with\_evidence & 0.464 & 0.964 & 0.964 & 0.964 \\
graph\_explanation\_hub    & 0.000 & 0.750 & 0.750 & 0.750 \\
transitive\_justification  & 0.846 & 0.885 & 0.962 & 0.962 \\
\bottomrule
\end{tabular}
\caption{Balanced accuracy by question type for the ensemble.}
\label{tab:ensemble_by_type}
\end{table*}

We evaluate four conditions: BL (baseline, no KG context), VO (oracle grounding using the relevant KG subgraph), PAT (grounding using hotspot pattern context derived from the simple-query evaluation), and LAY (layered selection over BL, VO, and PAT).

Scoring combines answer KG-alignment with edge-citation coverage. All results are reported using a majority-vote ensemble over the models (Table~\ref{tab:ensemble_results}).

Without graph grounding, ensemble balanced accuracy is only 0.515 despite relatively high raw accuracy (0.690), indicating that models often produce superficially plausible answers while failing to recover the KG-aligned semantic structure. The weakest category is graph explanation hub (0.000 balanced accuracy), showing that enumerating relations around a semantic hub is unreliable from parametric knowledge alone.

Injecting oracle KG grounding (VO) substantially improves performance, raising ensemble balanced accuracy to 0.900 and cross-model consistency from 0.868 to 0.948. More importantly, hotspot-pattern grounding (PAT) matches or exceeds VO on most models, increasing ensemble balanced accuracy to 0.903 and cross-model consistency to 0.974. This shows that the structural patterns discovered through simple equivalence-cluster evaluation transfer to more realistic compositional reasoning tasks.

The layered strategy (LAY) further improves reliability by selectively applying stronger grounding only where simpler grounding fails. LAY eliminates the 15 regressions introduced by PAT over already KG-consistent clusters, achieving 0.960 raw accuracy and 0.919 balanced accuracy, while reaching perfect performance on cycle-directionality queries.

The remaining limitation is graph explanation hub (0.750 balanced accuracy under LAY). Unlike local transitive or directional reasoning, these queries require exhaustive recall of a node's neighborhood rather than recovery of a small structural constraint. This suggests that localized hotspot grounding transfers effectively to compositional semantic reasoning, while full neighborhood reconstruction remains a harder retrieval problem that will remain a key task of our future work. 

\section{Complex queries (manual)}
\label{appendix:complex_queries}

\begin{description}

\item[Type:] Property inheritance \\
\textbf{Domain:} finance \\
\textbf{Q:} If bonds are used by investors, are municipal bonds also used by investors? \\
\textbf{A:} Yes \\
\textbf{Reasoning:} (municipal bond, subconcept of, bond), (bond, used by, investors)

\item[Type:] Property inheritance \\
\textbf{Domain:} finance \\
\textbf{Q:} If all insurance is taxed in my country, are Health Reimbursement Accounts taxed? \\
\textbf{A:} Yes \\
\textbf{Reasoning:} (Health Reimbursement Accounts, subconcept of, insurance)

\item[Type:] Property inheritance \\
\textbf{Domain:} finance \\
\textbf{Q:} Is money always a kind of legal tender? \\
\textbf{A:} No \\
\textbf{Reasoning:} (legal tender, subconcept of, money), NOT (money, subconcept of, legal tender)

\item[Type:] Hierarchy organization \\
\textbf{Domain:} finance \\
\textbf{Q:} What is the most generic concept, government bonds or loans? \\
\textbf{A:} loans \\
\textbf{Reasoning:} (government bond, subconcept of, loan)

\item[Type:] Property inheritance \\
\textbf{Domain:} medical\_occupations \\
\textbf{Q:} Does a pediatric neurosurgeon know about neurology? \\
\textbf{A:} Yes \\
\textbf{Reasoning:} (pediatric neurosurgeon, field of occupation, neurology)

\item[Type:] Property inheritance \\
\textbf{Domain:} medical\_occupations \\
\textbf{Q:} List the medical fields of occupation of a pediatric neurosurgeon. \\
\textbf{A:} Pediatrics, Surgery, Neurology

\item[Type:] Hierarchy organization \\
\textbf{Domain:} medical\_occupations \\
\textbf{Q:} What is more precise, saying that a pediatric neurosurgeon is a surgeon or a medical specialist? \\
\textbf{A:} Pediatric neurosurgeon is a surgeon \\
\textbf{Reasoning:} (pediatric neurosurgeon, subconcept of, surgeon), (surgeon, subconcept of, medical specialist)

\item[Type:] Property inheritance \\
\textbf{Domain:} medical\_occupations \\
\textbf{Q:} If all neurosurgeons require board certification, would pediatric neurosurgeons require it too? \\
\textbf{A:} Yes \\
\textbf{Reasoning:} (pediatric neurosurgeon, subconcept of, neurosurgeon)

\item[Type:] Property inheritance \\
\textbf{Domain:} medical\_occupations \\
\textbf{Q:} The insurance form only covers consultations with surgeons. Would an appointment with a pediatric neurosurgeon be eligible? \\
\textbf{A:} Yes \\
\textbf{Reasoning:} (pediatric neurosurgeon, subconcept of, neurosurgeon), (neurosurgeon, subconcept of, surgeon)

\item[Type:] Property inheritance \\
\textbf{Domain:} medical\_occupations \\
\textbf{Q:} Do all obstetrician gynecologists know about obstetrics? \\
\textbf{A:} Yes \\
\textbf{Reasoning:} (obstetrician gynecologist, field of occupation, obstetrics)

\item[Type:] Property inheritance \\
\textbf{Domain:} medical\_occupations \\
\textbf{Q:} Do all dentists know about oral medicine? \\
\textbf{A:} Yes \\
\textbf{Reasoning:} (dentist, field of occupation, oral medicine)

\item[Type:] Hierarchy organization \\
\textbf{Domain:} academic\_disciplines \\
\textbf{Q:} Organize the following concepts from most specific to most general: Theology, Christian Theology, academic discipline. \\
\textbf{A:} Christian Theology, Theology, academic discipline \\
\textbf{Reasoning:} (Christian Theology, subconcept of, Theology), (Theology, subconcept of, academic discipline)

\item[Type:] Hierarchy organization \\
\textbf{Domain:} dishes \\
\textbf{Q:} Which of these is a type of fish dish: Watermelon, sofrito, sashimi, tofu ramen? \\
\textbf{A:} Sashimi \\
\textbf{Reasoning:} (sashimi, subconcept of, fish dish)

\item[Type:] Hierarchy organization \\
\textbf{Domain:} music\_genres \\
\textbf{Q:} Can we say that liturgical music is the same as church music? \\
\textbf{A:} No \\
\textbf{Reasoning:} (church music, subconcept of, liturgical music), NOT (liturgical music, subconcept of, church music)

\item[Type:] Hierarchy organization \\
\textbf{Domain:} music\_genres \\
\textbf{Q:} Can we say that liturgical music is a kind of church music? \\
\textbf{A:} No \\
\textbf{Reasoning:} NOT (liturgical music, subconcept of, church music)

\item[Type:] Property inheritance \\
\textbf{Domain:} dishes \\
\textbf{Q:} Can all poultry dishes be considered meat dishes? \\
\textbf{A:} Yes \\
\textbf{Reasoning:} (poultry dish, subconcept of, meat dish)

\item[Type:] Hierarchy organization \\
\textbf{Domain:} dishes \\
\textbf{Q:} Can these three concepts be organized in a concept hierarchy: fried fish, dish, fish dish? \\
\textbf{A:} Yes \\
\textbf{Reasoning:} (fried fish, subconcept of, fish dish), (fish dish, subconcept of, dish)

\item[Type:] Hierarchy organization \\
\textbf{Domain:} dishes \\
\textbf{Q:} Can we say that fruit is a part of a watermelon salad? \\
\textbf{A:} Yes \\
\textbf{Reasoning:} (watermelon salad, has part, watermelon), (watermelon, subconcept of, fruit)

\item[Type:] Hierarchy organization \\
\textbf{Domain:} dishes \\
\textbf{Q:} If I say that all fruit is expensive, does it follow that watermelon salad is expensive? \\
\textbf{A:} Yes \\
\textbf{Reasoning:} (watermelon salad, has part, watermelon), (watermelon, subconcept of, fruit)

\item[Type:] Hierarchy organization \\
\textbf{Domain:} software \\
\textbf{Q:} If I say that operating systems are complex, does it follow that GNU/Linux is complex? \\
\textbf{A:} Yes \\
\textbf{Reasoning:} (GNU/Linux, subconcept of, Linux), (Linux, subconcept of, operating system)

\item[Type:] Hierarchy organization \\
\textbf{Domain:} software \\
\textbf{Q:} If I say that a firewall is a software appliance, does it follow that it is also an application? \\
\textbf{A:} Yes \\
\textbf{Reasoning:} (firewall, subconcept of, software appliance), (software appliance, subconcept of, application)

\item[Type:] Hierarchy organization \\
\textbf{Domain:} academic\_disciplines \\
\textbf{Q:} What is more general, pre-Columbian studies or Native American studies? \\
\textbf{A:} Native American studies \\
\textbf{Reasoning:} (pre-Columbian studies, subconcept of, Native American studies)

\item[Type:] Hierarchy organization \\
\textbf{Domain:} finance \\
\textbf{Q:} How do I file my statements for my health savings account, under insurance or under bonds? \\
\textbf{A:} Under insurance \\
\textbf{Reasoning:} (health savings account, subconcept of, insurance)

\item[Type:] Hierarchy organization \\
\textbf{Domain:} finance \\
\textbf{Q:} Which of the following products are insurance products: Health savings account, bond, flexible spending account, municipal bond, financial product? \\
\textbf{A:} Health savings account, flexible spending account \\
\textbf{Reasoning:} (health savings account, subconcept of, insurance), (flexible spending account, subconcept of, insurance)

\item[Type:] Hierarchy organization \\
\textbf{Domain:} academic\_disciplines \\
\textbf{Q:} Is it true that every book on Native American studies also covers pre-Columbian studies? \\
\textbf{A:} No \\
\textbf{Reasoning:} (pre-Columbian studies, subconcept of, Native American studies), NOT (Native American studies, subconcept of, pre-Columbian studies)

\end{description}

\section{Use of AI}
\label{appendix:use of AI}

Generative AI was used to do code refactoring, such as refactoring manually written code to parameterize models and datasets. 

In the writing of the paper itself, AI was used to create the latex tables from csv data as well as formatting SPARQL queries into the ACL double column format. Also, it was used to create the code that generates heats maps from csv data. Furthermore, individual paragraphs were fed into AI to improve clarity. 

All generative AI outputs were verified for quality and accuracy by the authors. 

\section{Background Datasets}
The MITRE D3FEND ontology~\citep{mitre_d3fend_ontology} uses a ``non-exclusive, royalty-free license to use D3FEND for research, development, and commercial purposes.'' The license can be viewed here: \href{https://d3fend.mitre.org/tou/}{https://d3fend.mitre.org/tou/}. The datasets on multiple Wikidata domains \citep{Uceda_2024} uses the \href{https://huggingface.co/datasets/choosealicense/licenses/blob/main/markdown/apache-2.0.md}{Apache 2.0 license}. Both datasets are intended for multiple uses, including research, which is consistent with our usage here. It is also consistent with the intended use of \textsc{SI-Cyber}. Both datasets are publicly available resources and refer to abstract concepts in different domains. MITRE D3FEND describes abstract categories related to cybersecurity while the Wikidata dataset describes abstract categories relates to the following: academic disciplines, dishes, finance, home appliances, medical specialties, music genres, natural disasters, and software. These datasets do not identify individual people or contain offensive content. Labels, comments, and metadata are represented in English.

\end{document}